\documentclass[sigconf]{acmart}

\usepackage{subcaption}
\setcopyright{none}
\acmISBN{}
\acmDOI{}
\copyrightyear{https://xai.kdd2019.a.intuit.com/}
\AtBeginDocument{%
  \providecommand\BibTeX{{%
    \normalfont B\kern-0.5em{\scshape i\kern-0.25em b}\kern-0.8em\TeX}}}

\usepackage{amsmath}
\usepackage{tabularx}

\newcommand{\BIT}{\begin{itemize}}
\newcommand{\EIT}{\end{itemize}}
\newcommand{\BNUM}{\begin{enumerate}}
\newcommand{\ENUM}{\end{enumerate}}

\newcommand\mbb[1]{\mathbb{#1}}

\def\mrm#1{\mathrm{#1}}
\def\indic#1{\mbb{I}\left({#1}\right)} 

\def\P{\mathbb{P}} 

\def\Var{\mrm{Var}}

\def\Varsubarg#1#2{\Var_{#1}\left(#2\right)}

\begin{document}

\title[Discovering Concepts in Learned Reresentations]{Discovering Concepts in Learned Representations using Statistical Inference and Interactive Visualization}

\author{Adrianna Janik}
\email{ada.janik@gmail.com}
\author{Kris Sankaran}
\email{kris.sankaran@umontreal.ca}

\affiliation{%
  \institution{Montreal Institute for Learning Algorithms}
  \streetaddress{6666 St-Urbain, #200}
  \city{Montr\'eal}
  \state{Qu\'ebec}
  \postcode{H2S 3H1}
}



\begin{abstract}
Concept discovery is one of the open problems in the interpretability literature
that is important for bridging the gap between non-deep learning experts and
model end-users. Among current formulations, concepts defines them by as a
direction in a learned representation space. This definition makes it possible
to evaluate whether a particular concept significantly influences classification
decisions for classes of interest. However, finding relevant concepts is
tedious, as representation spaces are high-dimensional and hard to navigate.
Current approaches include hand-crafting concept datasets and then converting
them to latent space directions; alternatively, the process can be automated by
clustering the latent space. In this study, we offer another two approaches to
guide user discovery of meaningful concepts, one based on multiple hypothesis
testing, and another on interactive visualization. We explore the potential
value and limitations of these approaches through simulation experiments and an
demo visual interface to real data. Overall, we find that these techniques offer
a promising strategy for discovering relevant concepts in settings where users
do not have predefined descriptions of them, but without completely automating
the process.
\end{abstract}



\maketitle

\section{Introduction}

Automatic representation learning can be both a blessing and a curse. On the one hand,
it allows practitioners to achieve state-of-the-art performance on a variety of
prediction tasks using a generic workflow, without having to hand-engineer
features from unstructured input, and it lies at the core of almost all modern AI
systems \cite{bengio2012unsupervised}. On the other, the resulting features can be difficult to characterize,
either intrinsically, or through their influence on downstream predictions.
Mitigating the risk that comes from deploying inscrutable systems in socially
relevant settings has emerged as a research priority in the machine learning community \cite{doshi2017accountability, samek2017explainable}.

\citet{kim_interpretability_2017} have proposed one way forwards, formally defining activation
scores for user-defined concepts. Their method, Testing with Concept Activation Vectors (TCAV), gives a score for how concepts --
practically, a collection of related images -- influence predictions. For
example, a user trying to understand a model for zebra classification can provide a "stripes" concept by collecting many images of stripes, and TCAV will generate a score describing the relevance of that concept. This offers a degree of
agency in model inspection not available in more traditional
interpretability techniques. The approach takes as its starting point an active
and curious user, willing to dedicate time to model inspection.

Our work takes this perspective one step further -- helping users become data
detectives -- in order to address one of TCAV's limitations, the need to collect
images from which concepts can be defined. Anything that can facilitate users'
interrogation of concepts has the potential to make the method more widely
applicable. In its focus on weakening the requirement for preset concepts, our work
is similar to \cite{ghorbani_automating_2019}. However, rather than automating concept discovery,
we aim instead to streamline it from the user's perspective, borrowing techniques from statistical inference and interactive visualization 
\cite{efron_large-scale_2010, wickham2015visualizing}. We have two main ideas,

\begin{itemize}
\item Multiple hypothesis testing: Rather than testing one well-defined
concept-related hypothesis at a time, we can sift through many (usually
uninteresting) candidates in search of a few promising ones, controlling for the
risk that indiscriminate searches produce false positives.
\item Interactive visualization: A visual interface can help 
users rapidly cycle through many concepts, offering feedback about queries on
the fly, and shaping intuition about relevant concepts in the process.
\end{itemize}

Both approaches suggest potential concepts in reference to existing samples, rather than
those collected explicitly for the purpose of concept evaluation. The
distinction is analogous to the one between exploratory and confirmatory data
analysis: the strategy outlined in this work allows weak identification across a
range of potential concepts, but not explicit verification of one defined in
advance.

Our main contributions area
\begin{itemize}
\item In Section \ref{sec:mht_method}, we develop a proposal for discovering
concepts using multiple hypothesis testing.
\item In Section \ref{sec:vis_method}, we describe a visual interface through
which users can interact with concept activation vectors.
\item In Sections \ref{sec:mht_illustration} and \ref{sec:vis_expr}, we experiment with and critique these proposals.
For multiple hypothesis testing, we leverage an illustrative simulated dataset, while
for interactivity, we consider a real-world computer vision task.
\end{itemize}

Further, our study takes a geometric perspective of concept
activation vectors; clarifying this interpretation is one of the implicit goals
of this work.

\section{Methodology}

Before detailing our proposals, we summarize the approach of \citet{kim_interpretability_2017}.
Say that there are $n$ samples $x_i$, each with corresponding label $y_i$. We
have a classifier $f$, mapping samples to predicted classes. $f$ can be an
arbitrary black box, with the exceptions that (1) we must be able to extract
automatically learned features corresponding to any input, and (2) it must
produce predicted probabilities $p_{k}\left(x_i\right)$ for each class $k$.

TCAV looks at how perturbations in the learned features -- call them $z_i$ -- affect
downstream predictions. The perturbation itself is derived from a user-specified
collection of images, $\left(x_i^{C}\right)$, which capture some concept $C$.
Specifically, each $x_i^{C}$ is featurized as $z_i^{C}$, and a classifier is
trained to linearly separate the $z_i^{C}$ from random $z_i$. The direction $v$
orthogonal to the separating hyperplane is used to summarize the concept $C$. By
changing $z_i$ to $z_i + \epsilon v_i$ for some small $\epsilon$, we
imagine we are perturbing sample $i$ slightly ``towards'' the concept $C$.

To measure the relevance of a concept with respect to the $K$ classes of the
response, we can calculate $S_{v}\left(x_i\right) = Dp\left(z_i\right)v$,
where $Dp\left(z\right)$ is the $K \times J$ dimensional Jacobian, whose
$kj^{th}$ entry, $\frac{\partial p_{k}\left(z\right)}{\partial z_{j}}$
describes the sensitivity of the $k^{th}$ class's probability with respect to
the $j^{th}$ learned feature, evaluated at the point $z$. This is a directional
derivative of the class probabilities in the direction $v$. TCAV quantifies the
significance of the concept encoded by $v$ with respect to the class $k$ by
seeing whether the fraction of $x_i$ from that class with positive directional
derivative is larger than would be expected when defining random concepts, where
a random concept is one defined by a random subset of the $x_i$.
Formally, the relevance of concept $v$ for class $k$ is given by the score
\begin{align}
\label{eq:tcavk}
TCAV_{k}\left(v\right) =\frac{|\left \{ x : S_{vk}(x) > 0 \text{ and } y_i = k \right \}|}{N_k}
\end{align}
where $N_k$ is the number of samples in class $k$ and $S_{vk}\left(x\right)$ 
is the $k^{th}$ coordinate of $S_v\left(x\right)$.

\subsection{Multiple Hypothesis Testing}
\label{sec:mht_method}

Multiple hypothesis testing is widely used in genetics, and we motivate its
application to the problem of concept discovery by drawing parallels to that
field. In genetics, when researchers want to know which genes are associated with
a particular disease, they perform a hypothesis tests for each of possibly
thousands of candidates, with the expectation that almost all will
be irrelevant. However, a few may emerge as potential signal within the noise,
and using an appropriate statistical correction, it is possible to guarantee
that the proportion of false discoveries among these selected candidates lies
below some fraction \cite{efron_large-scale_2010}.

In the concept discovery setting, we screen for concept vectors $v$ in the way
that a scientist screens for disease causing genes. To implement the idea, we
need to specify two things,

\begin{itemize}
\item A way of proposing concept directions to screen over.
\item A way of evaluating the significance of candidate concepts.
\end{itemize}

To propose concept directions, we simply draw $J$ random directions on the unit sphere in the learned
feature space. This has the disadvantage of undersampling directions with
high densities of $z_i$, but appears sufficient in our preliminary experiments.

Given this collection of directions, we could evaluate their individual
significance using a randomization test, as in \cite{kim_interpretability_2017}.
Rather than selecting random sets $\left(x_i^{C}\right)$ and fitting separating
hyperplanes, we pick $J^\prime$ random directions $v_j^{\ast}$ as above. This
has the benefit of not having to fit a separate hyperplane for each draw from
the randomization null, but again undersamples dense regions.

Given these reference $v_j^{\ast}$, a $p$-value can be obtained as\footnote{For clarity, we omit the class $k$ from our notation, but all these calculations must be done with respect to some set class.} $p_j =
\frac{1}{J^\prime}\left|\left\{TCAV\left(v_j^{\ast}\right) \geq
TCAV\left(v_j\right)\right\}\right|$, the tail area in the null reference distribution
of TCAV scores above the observed score for the candidate concept $v_j$ of interest. These $p$-values need to be
adjusted in order to guarantee proper statistical inference. To see why, note
that in the completely null case, where the $p$-values are uniformly
distributed, an $\alpha$ significance threshold for individual hypotheses
will lead to a fraction $\alpha$ of the $v_j$'s being declared interesting, when
really none of them are: 100\% of the discoveries are false discoveries. The
usual terminology is that the false discovery rate (FDR, the proportion of rejected
hypotheses that were in fact null) is 1.

To address this, we can use the Benjamini-Hochberg procedure: apply more
stringent thresholds, depending on strength of the signal \cite{efron_large-scale_2010}. Specifically, sort
the $p$-values, so that $p_{\left(j\right)}$ is the $j^{th}$ smallest of the
original $J$. Instead of comparing $p_{\left(j\right)} \leq \alpha$, compare
$p_{\left(j\right)} \leq \frac{j\alpha}{J} := \alpha_{j}$. Reject the hypotheses
corresponding to $p_{\left(1\right)}, \dots, p_{\left(j^{\ast}\right)}$, where $j^{\ast}$ is
last time the inequality $p_{\left(j\right)} \leq \alpha_{j}$ holds. This is
guaranteed to keep the proportion of ``false discovery'' directions $v_{j}$ below
$\alpha$.

This approach is the most natural one, but it suffers two limitations. First, it
requires a randomization test for each candidate. This is computationally
expensive, since samples cannot be reused -- this would violate the independence
assumption required to guarantee FDR control. Second, we observe that the
randomization $p$-values using the statistic \ref{eq:tcavk} can be degenerate. For
some directions, all $S_{vk}\left(x_i\right)$ are positive across all samples in
class $k$, so that the corresponding $p$-value must be zero.

We can address the first issue by using local FDR (lFDR) methods
\citep{efron_large-scale_2010}. Rather than starting with $p$-values, these methods operate
directly on test statistics. By assuming that most hypotheses are null, it is
able to estimate a reference null distribution. By comparing this reference with
the original test statistics, a local FDR can be approximated for any value of
the test statistic. Those in the bulk of the reference null have a high lFDR --
rejecting these would lead to a high fraction of false discoveries -- while those in the tails have a low
lFDR. Rejecting all points with lFDR $\leq \alpha$ is guaranteed to control the
FDR below $\alpha$.

To deal with the second issue, were consider test statistics that account for
the size of the score $S_v\left(x_i\right)$, not simply the sign. This
possibility is noted by \cite{kim_interpretability_2017}, but not pursued there.
In our experiments, we find that $\sqrt{\Varsubarg{\hat{\P}_{n}}{S_{v}\left(x_i\right)}}$,
the empirical standard deviation of concept activations, has a distribution more amenable
to lFDR estimation.

Aside from the main thread of concept discovery, we note that it is possible to adapt
the multiple testing view to the related problem of screening for subregions of
the feature space for which a prespecified concept direction is relevant. We can
cluster samples, evaluate concept activation scores within each, and compare to
reference distributions. This can be used to
identify samples along decision boundaries, where class probabilities are
changing rapidly.

\subsection{Interactive Exploration}
\label{sec:vis_method}

Concepts are directions in the learned features space. By making it possible for
users to interact with this space, we can facilitate concept discovery. There
are two challenges in implementing this idea. First, the dimensionality of the
learned space is typically very large. Second, users may become annoyed with the 
number of interactions necessary to find and understand interesting, tangible
concepts. To address these issues, we appeal to techniques from the
interactive data visualization literature: dimensionality reduction and linking
\citep{cook_and_buja}.

\begin{figure}
    \centering
    \includegraphics[width=0.45\textwidth]{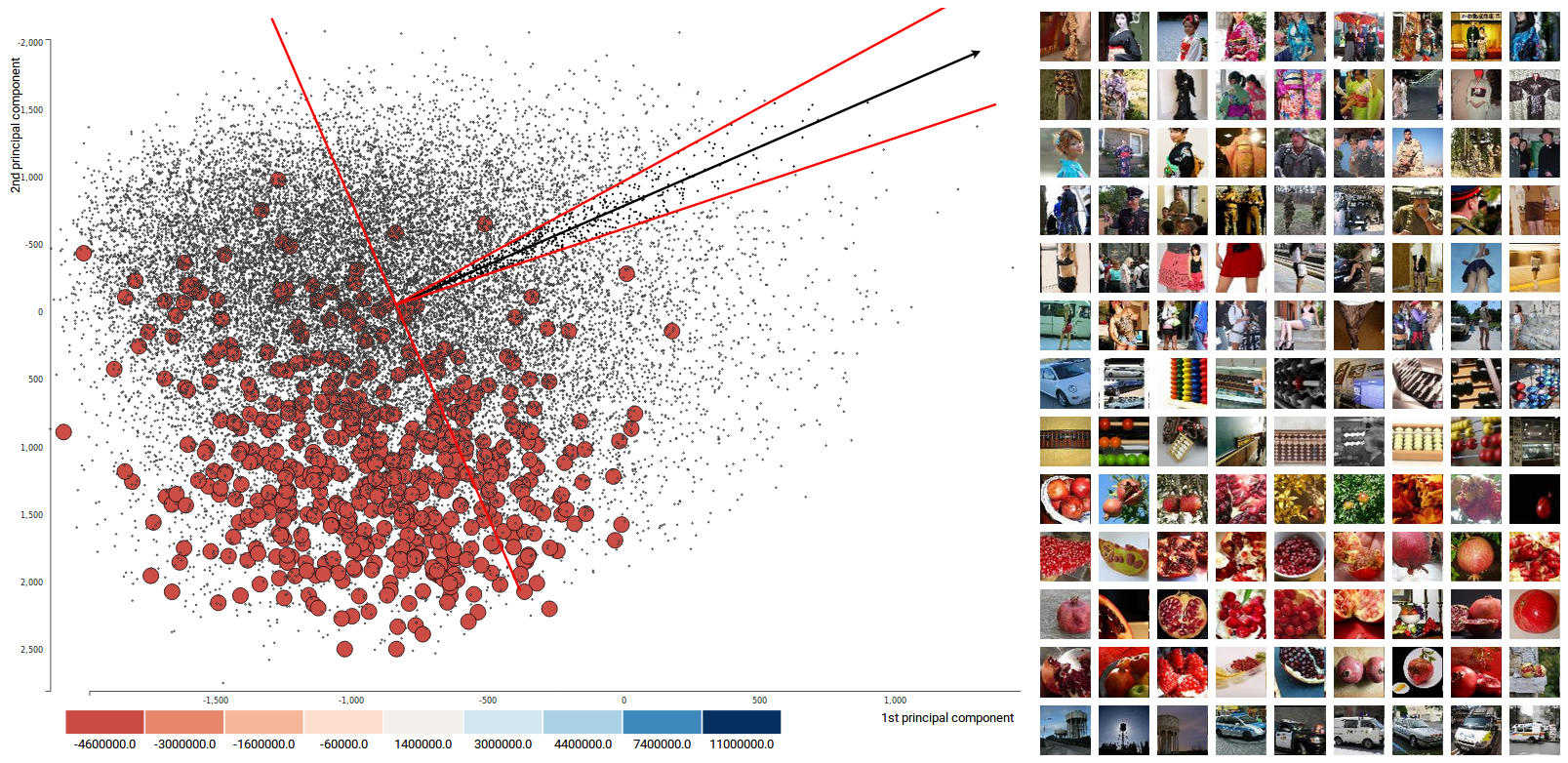}
    \caption{A screenshot from our demo, presenting a proof-of-concept for interactive concept discovery, available at
    \url{https://adrijanik.github.io/concepts-vis/}}
    \label{fig:annotated_interface}
\end{figure}

\begin{figure}
    \centering
    \includegraphics[width=0.45\textwidth]{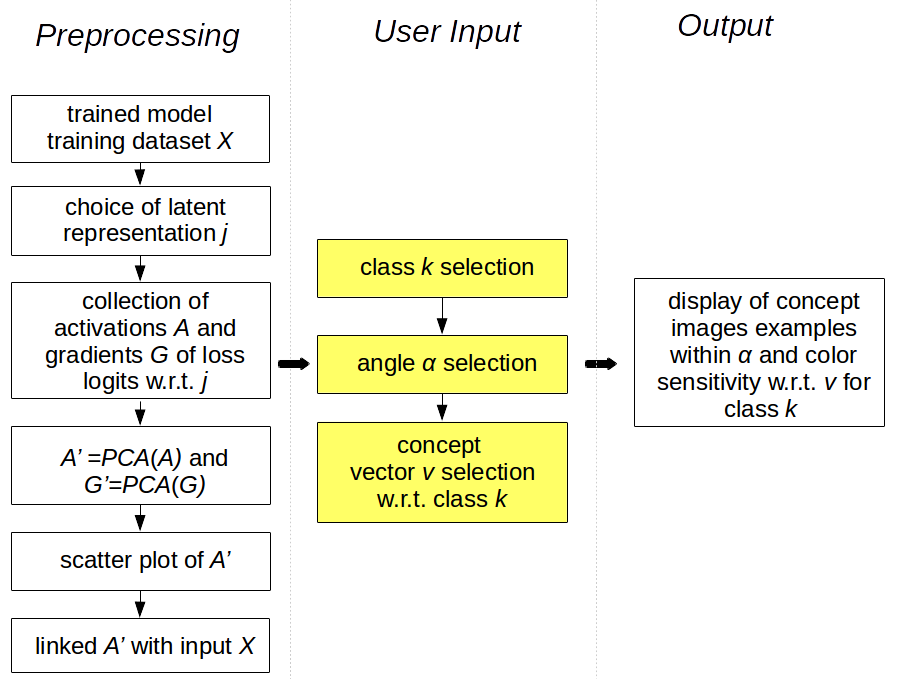}
    \caption{An overview of the procedure for interactively exploring concepts. The first column represents steps before interaction, the second column presents user interaction possibilities, and the third describes the expected output.}
    \label{fig:interface_diagram}
\end{figure}

\begin{figure}
    \centering
    \includegraphics[width=0.5\textwidth]{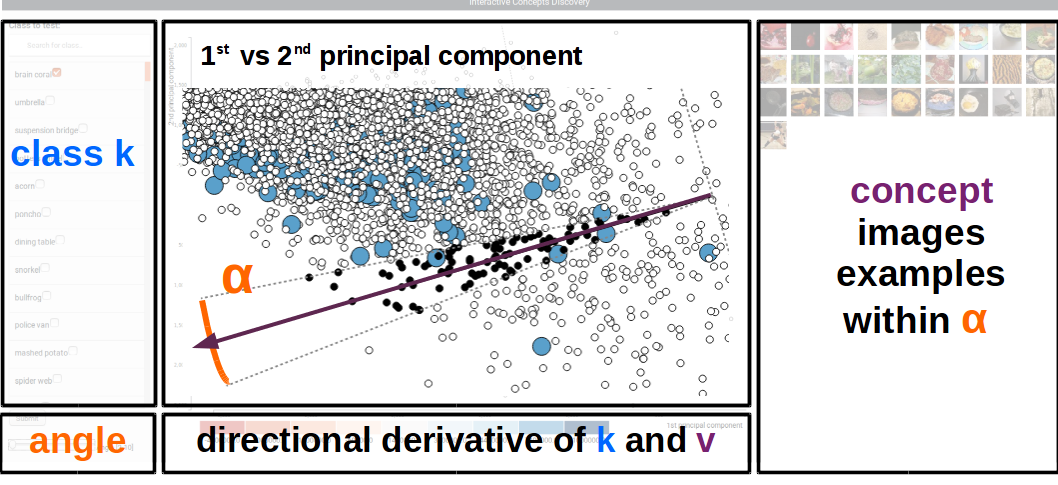}
    \caption{An outline of the key elements of our visualization interface.}
    \label{fig:interface-outline}
\end{figure}

We propose the view in Figure \ref{fig:annotated_interface}, with functionality outlined in Figure \ref{fig:interface-outline}. Each point in the
scatterplot corresponds to one sample in the original data. Their positions are
obtained by plotting with respect to the directions derived from a Principal
Components Analysis (PCA) on the learned features. This reduction is how we manage
the high-dimensionality of the $z_i$; alternative reductions could be used. The
central black arrow represents a user-defined concept direction $v$, and points of 
the currently selected class $k$ are enlarged and colored according to the class' score, $TCAV_{k}\left(v\right)$. This immediate visual feedback
on the relevance of a concept direction is how we make it easier to discover
interesting concepts. To provide context about the relevant directions, we
display representative images along the concept direction, between the dashed helper lines defined by a user-chosen angle in the bottom left corner. The left panel lets users switch between classes. The overall procedure is summarized in Figure \ref{fig:interface_diagram} -- the elements requiring user interaction are highlighted in yellow.

Working in lower dimensions has its cost, it is an approximation. To calculate the TCAV score, we need to obtain a directional derivative between the concept vector and the gradient of logit function with respect learned features. For the type of high-dimensional $z_i$ learned by modern deep learning models, performing this computation with the full high-dimensional gradient for each point breaks the system's responsiveness. Instead, we again apply dimensionality reduction, this time to the gradients. This trick allows us to store only projection of the gradients, dramatically decreasing the computational load.

The score obtained from the directional derivative is plotted in color space on samples from the selected class. In this way, users can quickly assess which directions are worth exploring further. Instead of staying in the projected space 
we could perform the reverse operation to obtain an approximation of the vector in the original feature space. In the case of PCA, we use the following procedure: let $X$ be the original $m \times n$ input matrix, $\mu$ be the mean vector of $X$ and let $V$ be the $n \times k$ matrix of eigenvectors with highest eigenvalues used to obtain the PCA scores $Z$. Then, the reconstruction in the original space is $\hat{X} = ZV^T  + \mu$. While this is possible, we find that it is still too slow for interactive visualization. 

To further streamline the search, we can imagine a few modifications. To suggest
directions worth exploring, we could precompute activation summary statistics
across directions, and provide reference to them on the plot. We could show the
loss evaluated on the projected space as a heatmap, highlighting ridges at which
class predictions change. We could also sort classes so that those with similar
active concept vectors are placed nearer to one another.

\section{Experiments}

We now present experiments to evaluate our proposed approaches, with the goal of
discovering strengths and clarifying weaknesses.

\subsection{Multiple Testing Illustration}\label{sec:mht_illustration}

\begin{figure}
    \centering
    \includegraphics[width=0.5\textwidth]{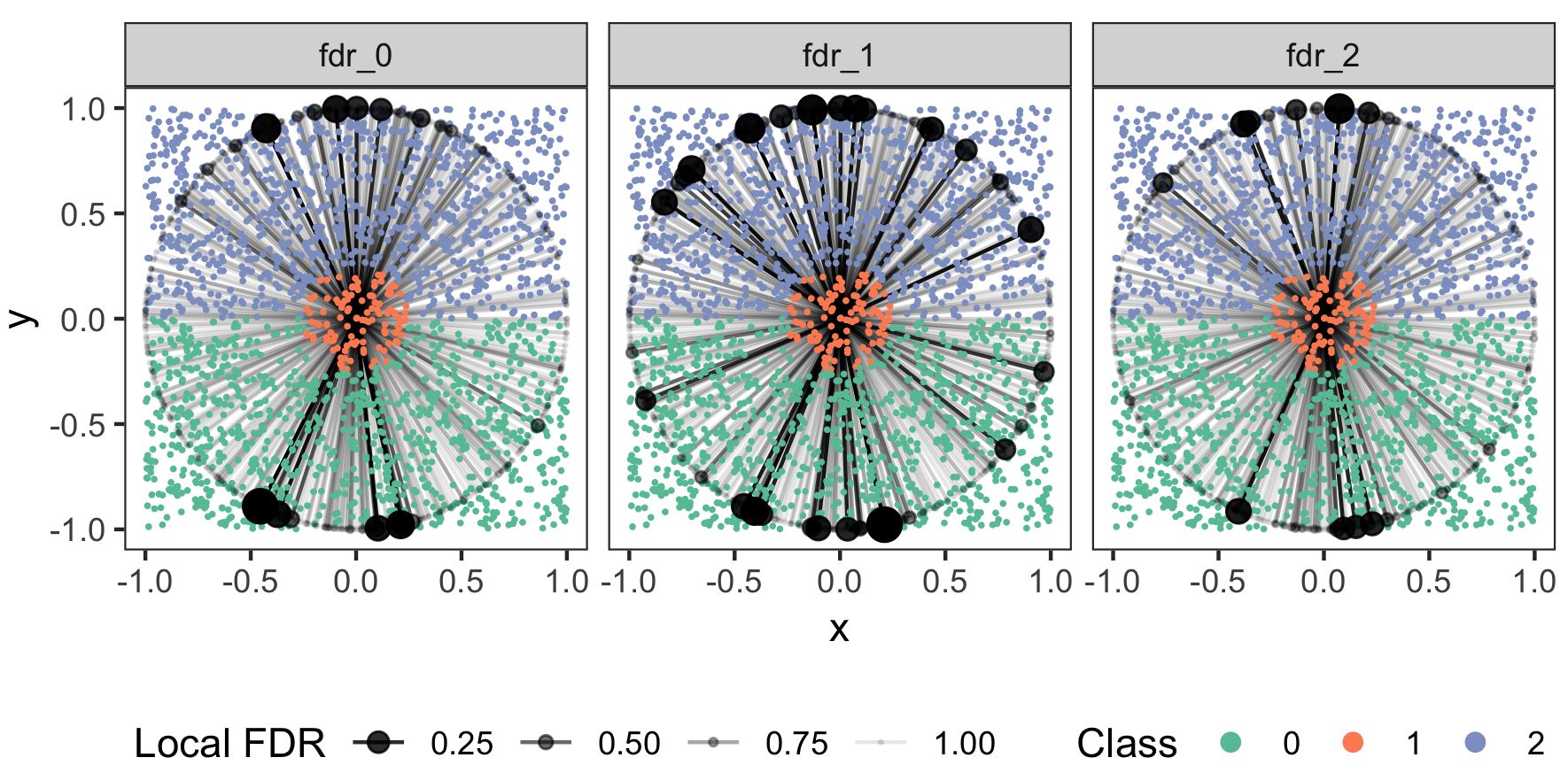}
    \caption{Class 1 has discovered (low lFDR) concepts in a variety of directions, while the low lFDR directions for classes 0 and 2 are generally oriented vertically.}
    \label{fig:fdr_directions}
\end{figure}

We first study the behavior of the multiple testing proposal in a transparent
simulation setting. The data are generated as follows. Draw 2000 points
uniformly in the square $\left[-1, 1\right] \times \left[-1, 1\right]$. 
Define three classes (see the background in Figure \ref{fig:fdr_directions}),

\begin{itemize}
\item Class 0 are points in $\{\left(x_1, x_2\right) : x_{2} < 0 \cap \|x\|_2 > 0.25 \}$.
\item Class 1 are points in $\{\left(x_1, x_2\right) : \|x\|_2 \leq 0.25 \}$.
\item Class 2 are points in $\{\left(x_1, x_2\right) : x_{2} \geq 0 \cap \|x\|_2 > 0.25 \}$.
\end{itemize}

The interesting characteristic of this dataset is that a large portion of the
decision boundaries for classes 0 and 2 is linear, while for class 1, the
boundary is a circle. Intuitively, the vertical direction is more important for
defining classes 0 and 2 than it is for class 1.

Our ``black box'' classifier is in fact a simple 1-layer MLP. That is,
the probability that $x$ belongs to class $k$ is approximated by
\begin{align*}
p_{k}\left(x\right) &= \left[\sigma\left(W^{(2)} z\left(x\right) + b^{(2)}\right)\right]_{k} \\
z\left(x\right) &= \text{ReLU}\left(W^{\left(1\right)} x + b^{(1)}\right),
\end{align*}
where $\sigma$ is the $K$-dimensional softmax and ReLU denotes the function $x \circ \indic{x > 0}$.
We use only 20 hidden units $z\left(x\right)$ -- this is enough to reach essentially 0
training loss. This model is convenient for our study of interpreting
automatically learned features because its learned mapping $z$ has a simple
description: each coordinate $z_{ik}\left(x_i\right) = \text{ReLU}\left(w^{(1)T}_{k}x_{i} + b^{(1)}_{k}\right)$ activates when $x_i$ lands in
the half-space orthogonal to $w^{(1)}_{k}$, with larger activations further in. By
mixing these activating hyperplanes, the MLP can trace out the curve defining
the true boundary. See Supplementary Figure \ref{fig:activations_with_surface}.

Given this setup, we can now ask,
\begin{itemize}
\item What concepts are discovered by the multiple testing procedure?
\item For any specific direction, which clusters of points are most strongly activated?
\end{itemize}

Considering the variants of the multiple hypothesis testing approach presented
in Section \ref{sec:mht_method}, we first describe our specific
implementation\footnote{All code for this study are available at
\url{https://github.com/adrijanik/concepts-exploration}, see Appendix \ref{sec:reproducibility} for more on reproducing experiments}. We draw 500 candidate
concepts $v_j$ on the 20-dimensional unit sphere in the learned feature space.
For fixed $v_j$ and class $k$, we evaluate $S_{v_j k}\left(x_{i}\right)$ over
all 2000 $x_i$.

\begin{figure}
    \centering
    \includegraphics[width=0.4\textwidth]{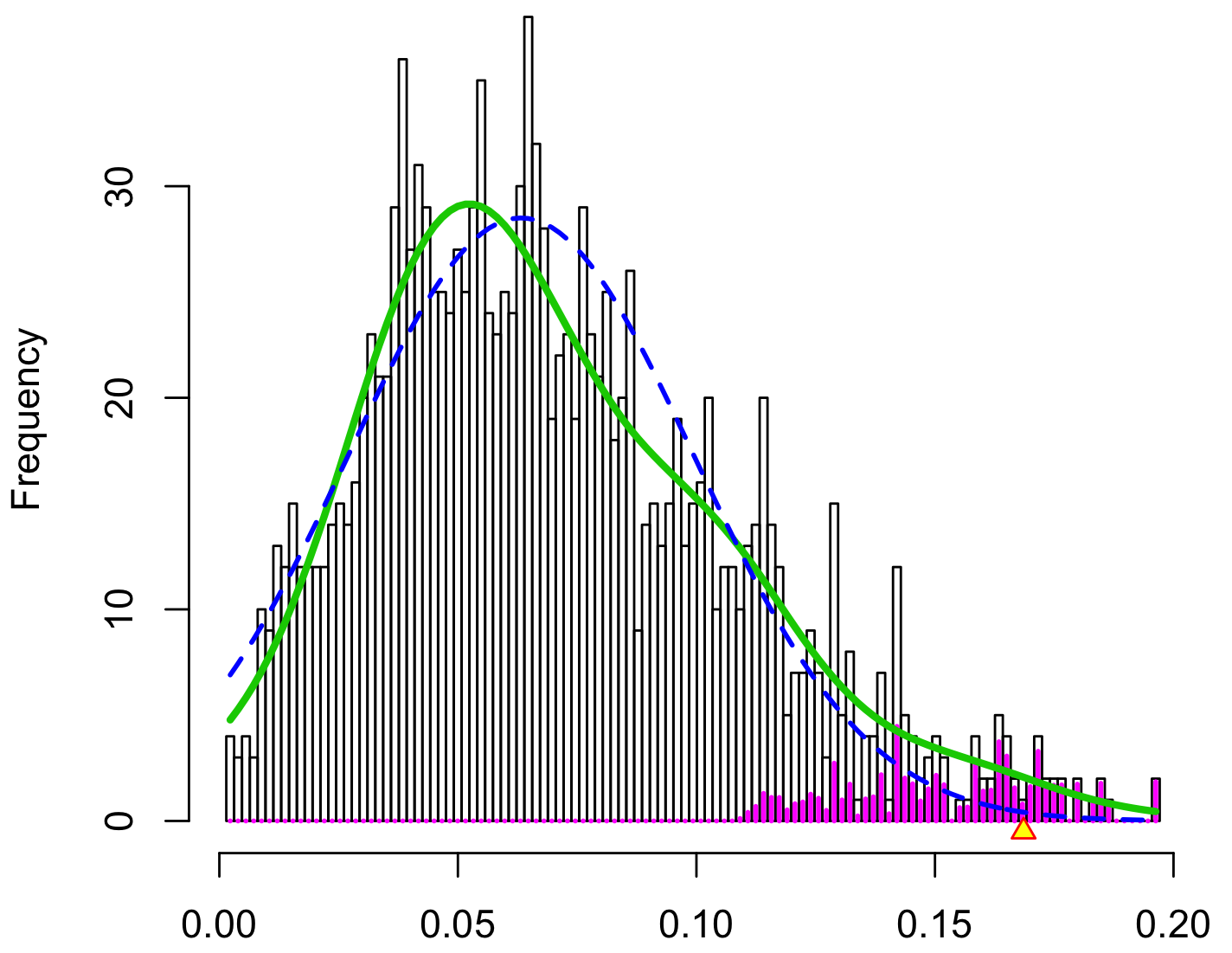}
    \caption{The Local FDR procedure fits the blue distribution as a reference null, which leads to the pink directions being assigned low lFDRs and flagged as interesting.}
    \label{fig:fdr_histogram}
\end{figure}

As a measure of interestingness, we compute the standard deviation
$\hat{\sigma}_{jk}$ of the collection
$\left(S_{v_{j}k}\left(x_i\right)\right)_{i = 1}^{2000}$. We use the local
FDR procedure applied to these $\hat{\sigma}_j$ to assign an lFDR scores to
every candidate concept. The reference null and declared discoveries are
displayed in Figure \ref{fig:fdr_histogram}.

Since our simulation data are low-dimensional, we can visualize the discovered
concepts, though only indirectly. To do so, recall that the $k^{th}$ learned
feature corresponds to an activation in the direction orthogonal to $w_k^{(1)}$.
Identify this direction with the $j^{th}$ standard basis vectors, $e_j$, in the
learned feature space. Formally, the duality comes from $W^{(1)T} e_j = w_j$;
informally, moving all points in the $w_j$ direction in the $x$-space increases
the activation of the $k^{th}$ learned features.

In this light, it is natural to associated the direction $W^{(1)T} v_j$ in the
$x$-space with the candidate concept $v_j$. These directions, normalized to
length 1, are displayed in Figure \ref{fig:fdr_directions}, shaded in according
to the lFDR of the associated direction, and overlaid on the original simulation
data. Evidently, the relevant directions for classes 0 and 2 generally lie
orthogonal to the $x$-axis. On the other hand, class 1 has low lFDR concept
directions that make relatively small angles with the $x$-axis. The fact that
the discovered directions correspond to those defining the decision boundaries
relevant to each class provides qualitative validation of our approach, though
our setting is admittedly a simple one.
    
\begin{figure}
    \centering
    \includegraphics[width=0.5\textwidth]{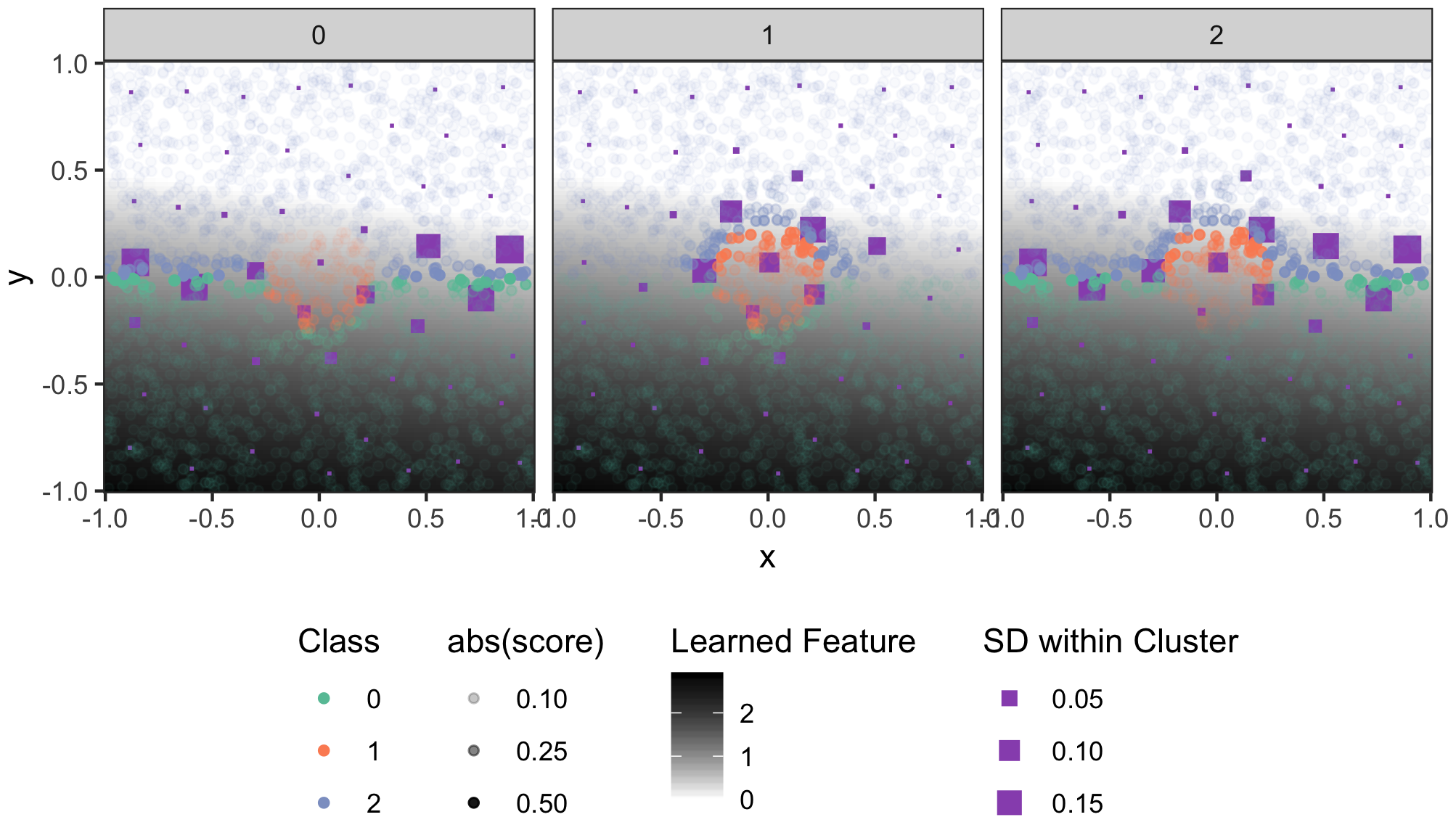}
    \caption{Concepts activations tend to have high variation near decision boundaries.}
    \label{fig:cluster_activation_sds}
\end{figure}

We next consider the distribution of $S_v\left(x_i\right)$ when a specific
direction is chosen, with the goal of identifying strongly activated subsets of
points. In a real application, this might provide a more granular view into
concept activations than the class-level summary statistic in equation
\ref{eq:tcavk}. To this end, we set the concept to $v = e_1$, corresponding to
the first learned feature dimension. We cluster all samples in the original
$x$-space using k-means and compute the standard deviation of activation scores
within each cluster. The result is displayed in Figure
\ref{fig:cluster_activation_sds}.

Each purple square corresponds to one cluster centroid. Their sizes reflects the
standard deviation of activations within that cluster, where activations are
computed with respect to the class $k$ indicated by the subpanel titles. The
absolute value of the original per-point activation scores is indicated by the
transparency of the points. The value of the learned feature
$z_{1}\left(x\right)$ is displayed as the grey plane in the background.

Clusters with the largest variation in concept activation for this direction
tend to appear near the decision boundaries for the corresopnding classes. Note
further that parts of the boundary that are far from the learned feature's
defining hyperplane tend to have smaller variation -- this is visible in the
U-shaped part of the boundaries for the first and second classes, for example.
In Supplementary Figure \ref{fig:cluster_by_activation}, we further find that
activations are generally positive for classes 0 and 1 and negative for class 2.
This is consistent with the downward orientation of the feature under study.

If this pattern generalizes outside of this simulation experiment, then
this approach to summarizing clusters by concept activation can be used to
sample points near decision boundaries where specific concepts -- user-defined
or statistically discovered -- are informative.
    
\subsection{Interactive Visualization}\label{sec:vis_expr}
For visualization, we experiment with a real-world image classification task. We inspect a GoogleNet network trained on ImageNet images. For the purpose of visualization we chose the 50 first classes from the 200 classes in tiny ImageNet. For the selected classes we obtain feature activations and gradients at the \texttt{mixed4d} layer. Our setup mirrors that in the experiments of \citet{kim_interpretability_2017}. For each class we have 500 images; in total, our dataset has 25000 images.  

\begin{figure}
    \centering
    \includegraphics[width=0.4\textwidth]{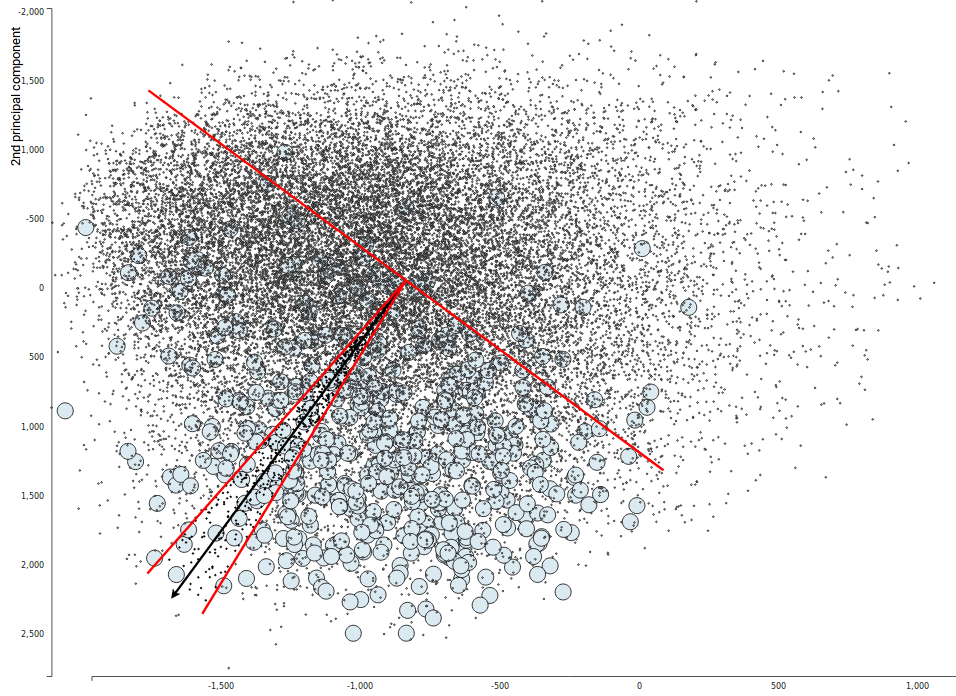}
    \caption{Concept with a high score for class ``flagpole'' (blue color), suggests that this concept is relevant for tested class.}
    \label{fig:relevant}
\end{figure}

\begin{figure}
    \centering
    \includegraphics[width=0.4\textwidth]{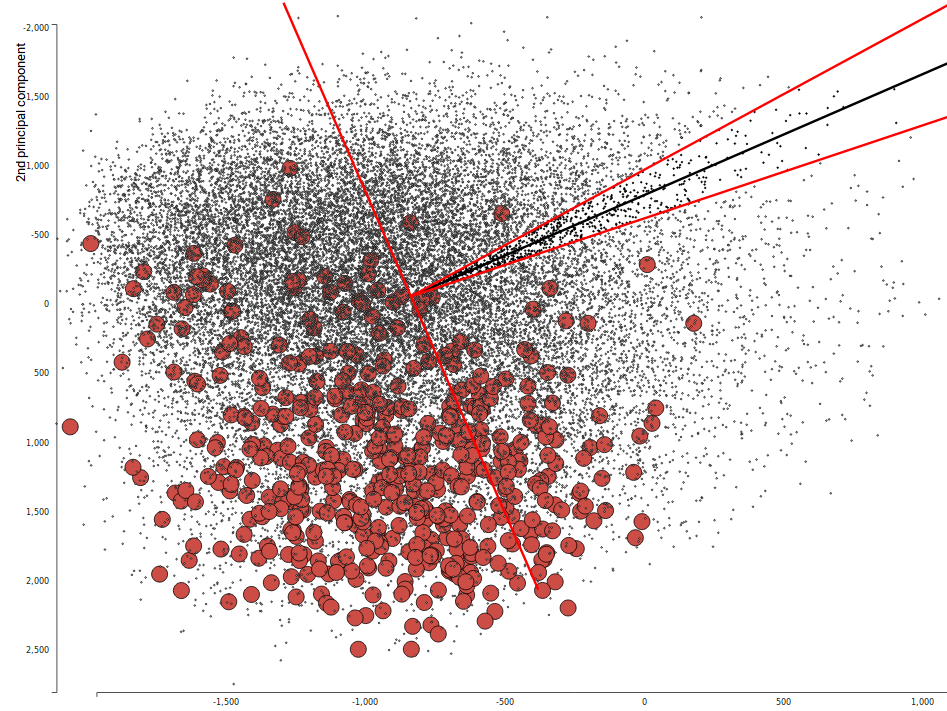}
    \caption{Concept with a very low score for class ``flagpole'' (red color), suggests that this concept is irrelevant for tested class.}
    \label{fig:irrelevant}
\end{figure}

\begin{figure}
\centering
\begin{minipage}{.5\textwidth}
  \centering
\begin{subfigure}{.48\textwidth}
  \includegraphics[width=0.95\textwidth]{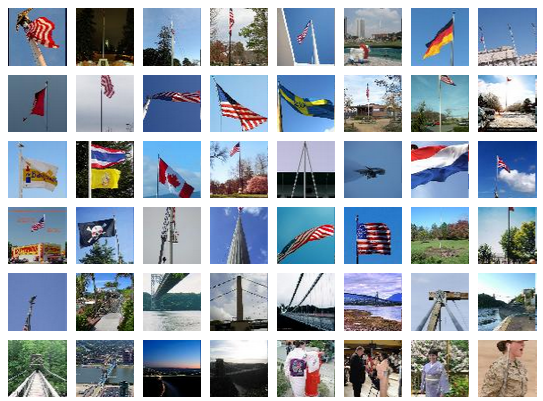}
  \caption{Relevant concept images \\ for "flagpole" class.}
  \label{fig:sub3}
\end{subfigure}%
\begin{subfigure}{.48\textwidth}
  \includegraphics[width=0.95\textwidth]{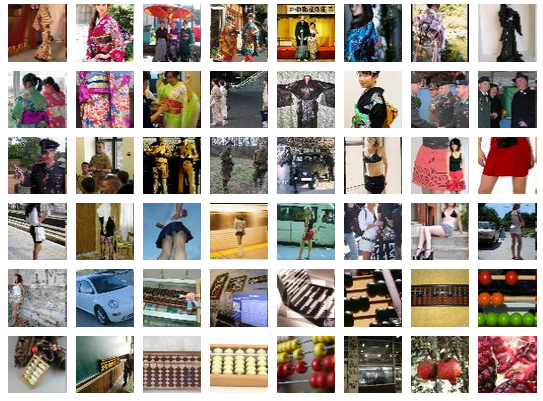}
  \caption{Irrelevant concept images \\ for "flagpole" class.}
  \label{fig:sub4}
\end{subfigure}
\caption{Images examples of relevant and irrelevant concepts from Figures \ref{fig:relevant} and \ref{fig:irrelevant}. We can see that among conceptually similar images to flagpole are masts and bridge support pillars, irrelevant images on the other hand seems not alike at all.}
\label{fig:examples}
\end{minipage}%
\end{figure}

In Figure \ref{fig:relevant}, we see an example of a user chosen concept vector pointing towards samples from the selected class ``flagpole.'' This is manifested by changes in the colors of points from that class -- points become dark blue when the user's concept vector leads to positive activation scores. In contrast, we can identify a direction that triggers negative scores, marked as red and presented in Figure \ref{fig:irrelevant}. Example concepts images can be found in Figure \ref{fig:examples}.

\begin{figure}
    \centering
    \includegraphics[width=0.4\textwidth]{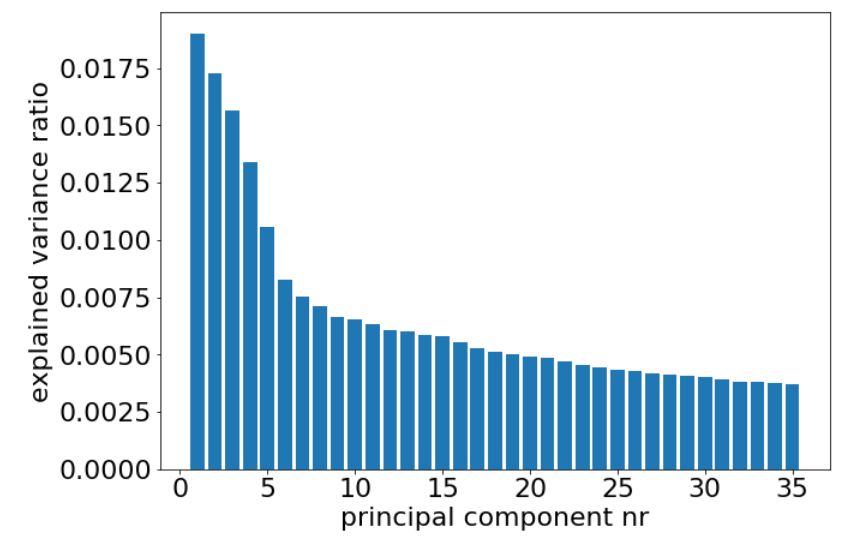}
    \caption{Bar chart of explained variance ratio for 35 first principal components. We can see that majority of components explain very little variance, but also first components variance ratio is below 2\% which rises concerns about plotting only first two components.}
    \label{fig:variance}
\end{figure}

It is worth questioning whether the dimensionality-reduced view is a reliable representation of original space. For this purpose, we look at the variance ratio explained by selected components. With a complex network such as GoogleNet, the activation space has dimension of $19\times 19 \times 528$ for each analyzed image. After flattening, this gives an activation vector of size $190,608$, and when we reduce this dimension with PCA, variance is distributed across many components (see Figure \ref{fig:variance}) with first one having around 2\% of variance explained. 

Visualizing CAVs in the lower dimension is an approximate approach but it may be useful. Navigating through latent space with hundred of thousands dimension is not an easy task and this initial experiments showed us that we may need to visualize more principal components in coordinated views or try our method with a smaller networks, especially promising may be networks with encoder-decoder structure due to their architecture and clear bottleneck layer.

We have potential ideas for future experiments we especially find promising plotting pairs of principal components in coordinated views to see how concept directions affect other sources of variance and also trying different methods of dimensionality reduction especially non-linear ones.

\section{Discussion}

Concept discovery can be guided by geometry, statistical inference, and
visualization. Inspired by \cite{kim_interpretability_2017}, we have developed
strategies that let practitioners interrogate the decision surfaces of deep
learning models and the way they relate to automatically learned features. We
intend for this approach to be a compromise between the manual concept
definition of \cite{kim_interpretability_2017} and the full automation of
\cite{ghorbani_automating_2019}.

There are important challenges that must be overcome before either
inferential or interactive concept exploration can become commonplace. The first
challenge we highlight is that there are likely more efficient strategies for
sampling candidate concepts $v_j$ in the learned feature space. It seems natural
to place more mass on directions containing more observations. Second, there is
the challenge of choosing an measure of concept interestingness which is
amenable to local FDR methodology, and a systematic study of
potential test statistics beyond those described here would be valuable. Third, our interactive
visualization approach is limited by difficulties of visualization
high-dimensional functions. Further experiments -- perhaps using parallel
coordinates or linked views \cite{cook_and_buja} -- may yield improvements.

Nonetheless, our experiments point to the value of model interpretability workflows that
equip users with all the tools of modern statistics and visualization. We have
provided experiments that demonstrate how multiple hypothesis testing can
highlight relevant concepts to follow-up on even when none are provided upfront.
Further, we have highlighted how interactive views that link concrete image
displays with abstract class summaries can engage users in the process of concept
discovery. We hope that these techniques play a role in advancing model
interpretability as a scientific enterprise, providing powerful tools to users
curious about the inner workings of complex models.

\bibliographystyle{ACM-Reference-Format}
\bibliography{ref}
\appendix

\section{Reproducibility}
\label{sec:reproducibility}

Demo was developed in JavaScript with the usage of d3.js visualization library. Demo is available online under following link: \url{https://adrijanik.github.io/concepts-vis/}.
Network that was been used is GoogleNet that was trained on ImageNet and used in the TCAV implementation tutorial by tensorflow. As the original training set is huge, we limited this study to a smaller dataset tiny-ImageNet-200 \url{https://tiny-imagenet.herokuapp.com/} with 200 classes from original set.

Pre-processing part of data was done using Python 3.6 with implementation of PCA from scikit-learn package. PCA was trained on 600 samples that contained 3 samples from each class drawn randomly from the dataset.

The source code can be found in the following GitHub repository: \url{https://github.com/adrijanik/concepts-exploration}.

\section{Supplementary Figures}

\begin{figure}[!hb]
    \centering
    \includegraphics[width=0.45\textwidth]{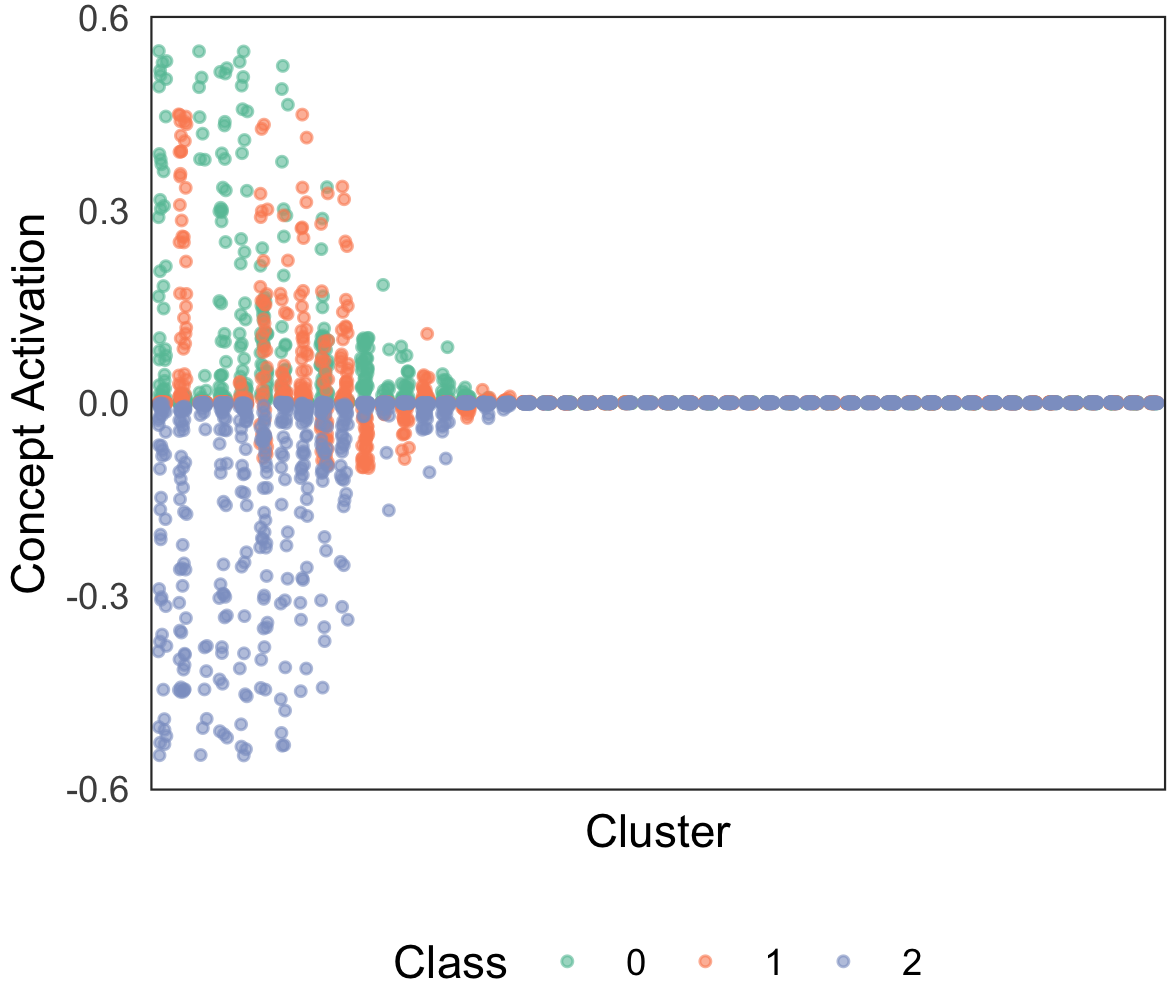}
    \label{fig:cluster_by_activation}
    \caption{We can identify subsets of samples with widely varying concept
    activation scores -- these tend to occur near decision boundaries. Each column
    of points corresponds to one cluster. Clusters have been sorted from those with
    largest variation in concept activation scores, to those with the least.
    Individual points are samples lying within those clusters, their vertical
    position is their score $S_{k}\left(v\right)$ for the $v$ presented in Figure
    \ref{fig:cluster_activation_sds}. Activations for different classes are colored
    separately. Evidently, in this concept direction, the probability surface
    between classes 0 and 1 is steeper than that between 1 and 2.}
\end{figure}    

\begin{figure}[!hb]
    \centering
    \includegraphics[width=0.45\textwidth]{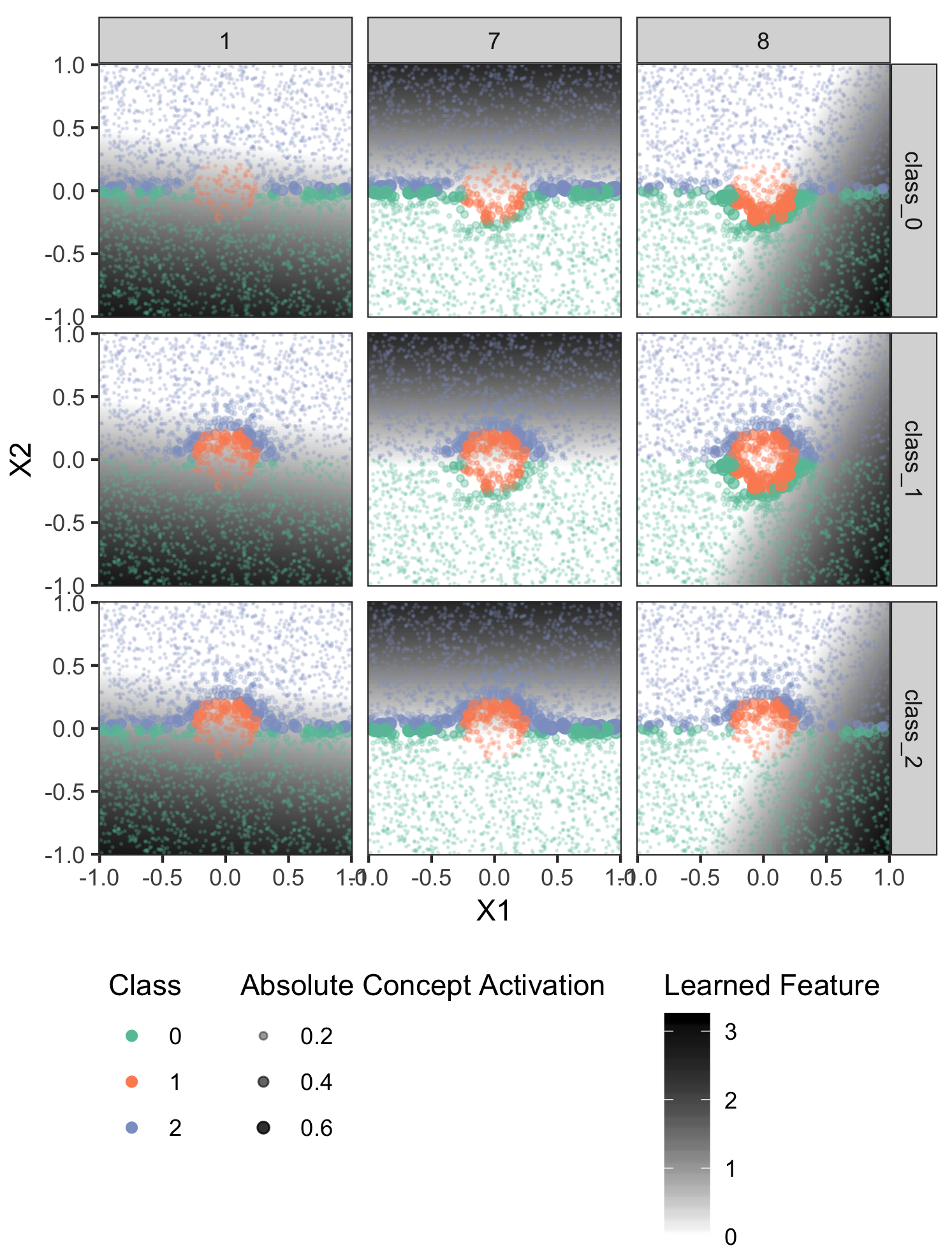}
    \label{fig:activations_with_surface}
    \caption{Simulation data overlaid with three learned features and concept activation scores. Each panel column corresponds to one of 20 learned features, which are represented through their activations as black and white half-spaces. Each panel row corresponds to a class on which the activation scores $S_{vk}\left(x\right)$ are displayed, where the direction $v = e_j$ for the associated feature coordinate $z_j$. The size and transparency of the points reflects their absolute concept score. Note that concept scores tend to be highest near the relevant class boundaries.}
\end{figure}    

\end{document}